\documentclass[letterpaper, 10 pt, conference]{ieeeconf}  % Comment this line out if you need a4paper

\IEEEoverridecommandlockouts                              % This command is only needed if 
\usepackage{xcolor}
\usepackage{amsfonts}       % blackboard math symbols
\usepackage{amsmath}
\usepackage[ruled]{algorithm2e}
\usepackage{bbm}
\usepackage{multirow}
\usepackage{amssymb}
\usepackage{fancyhdr}       % header
\usepackage{graphicx}       % graphics
\usepackage{tablefootnote}
\usepackage{bm}
\usepackage{hyperref}
\usepackage{dsfont}

\title{\LARGE \bf
Attention-based Proposals Refinement for 3D Object Detection
}

\author{Minh-Quan Dao, Elwan Héry, Vincent Frémont% <-this % stops a space
\thanks{Authors are with Nantes Université, École Centrale de Nantes and CNRS LS2N, 44300 Nantes, France
        {\tt\small first-name.last-name@ec-nantes.fr}}%
}

\begin{document}

\maketitle
\thispagestyle{empty}
\pagestyle{empty}

%%%%%%%%%%%%%%%%%%%%%%%%%%%%%%%%%%%%%%%%%%%%%%%%%%%%%%%%%%%%%%%%%%%%%%%%%%%%%%%%
\begin{abstract}

%Safe autonomous driving technology heavily depends on accurate 3D object detection since it produces input to safety critical downstream tasks such as prediction and navigation. 
Recent advances in 3D object detection are made by developing the refinement stage for voxel-based Region Proposal Networks (RPN) to better strike the balance between accuracy and efficiency. A popular approach among state-of-the-art frameworks is to divide proposals, or Regions of Interest (ROI), into grids and extract features for each grid location before synthesizing them to form ROI features. While achieving impressive performances, such an approach involves several hand-crafted components (e.g. grid sampling, set abstraction) which requires expert knowledge to be tuned correctly. This paper proposes a data-driven approach to ROI feature computing named APRO3D-Net which consists of a voxel-based RPN and a refinement stage made of Vector Attention. Unlike the original multi-head attention, Vector Attention assigns different weights to different channels within a point feature, thus being able to capture a more sophisticated relation between pooled points and ROI. Our method achieves a competitive performance of 84.85 AP for class Car at moderate difficulty on validation set of KITTI and 47.03 mAP (average over 10 classes) on NuScenes while having the least parameters compared to closely related methods and attaining an inference speed at 15 FPS on NVIDIA V100 GPU. The code is released~\url{https://github.com/quan-dao/APRO3D-Net}.

%This paper takes a more data-driven approach to ROI feature extraction using the attention mechanism. Specifically, points inside a ROI are positionally encoded to incorporate ROI 's geometry. The resulted position encoding and their features are transformed into ROI feature via Vector Attention. 

% Most state-of-the-art 3D detection frameworks are based on elaborated grid-based pooling to extract ROI (region of interest) features to retain as much as possible 3D structure in the refinement stage

% we can do so without grid pooling by rather let ROI where it want to draw information

% This paper develops a refinement stage for voxel-based object detection methods. Experiments on KITTI val set show that our method achieves competitive performance of 84.84 AP for class Car at Moderate difficulty while having the least parameters compared to closely related methods.

\end{abstract}

%%%%%%%%%%%%%%%%%%%%%%%%%%%%%%%%%%%%%%%%%%%%%%%%%%%%%%%%%%%%%%%%%%%%%%%%%%%%%%%%
\section{introduction} \label{sec:intro}
Object detection is a crucial component of autonomous vehicles because it provides input for downstream tasks such as the prediction of other road users' motion which essentially influence the motion planning of the ego vehicle. Due to the need for localizing objects in 3D space, object detection for autonomous vehicles is often performed on point clouds collected by 3D LiDAR. The unstructured and sparse nature of point clouds makes them unsuitable for convolutional neural networks (CNNs) to operate. Early works \cite{chen2017mv3d, yang2018pixor} rasterize point cloud to Bird-Eye View (BEV) to enable the use of standard 2D CNNs. Their encouraging results motivate studies on learning BEV representation of point cloud \cite{zhou2018voxelnet, yan2018second, lang2019pointpillars}. Their common point is the discretization of point clouds to 3D grids made of voxels, with PointPillars \cite{lang2019pointpillars} being the extreme case where voxels have infinite size along the vertical direction. Voxel-based methods have excellent inference speed thanks to the regular grid structure brought by the voxelization step. However, their performances are limited due to the lack of 3D structure in the BEV representation. 

Aware of such limitation, a number of works, e.g., \cite{shi2019pointrcnn, yang2019std, yang20203dssd}, advocate for operating directly on raw point cloud by using Set Abstraction and Feature Propagation (proposed by PointNet++ \cite{qi2017pointnet++}) instead of Convolution. The fine grain structure preserved by operating at the point level helps point-based methods outperform voxel-based methods in various benchmarks. The drawback of point-based methods is their low frame rate caused by point cloud query operators.

Recently, there has been a renaissance in voxel-based methods. This stems from the observation that voxel-based methods such as SECOND \cite{yan2018second} have exceptionally high recall rate (up to $95\%$) yet only achieves a moderate performance, e.g. SECOND's 78 Average Precision (AP) for Car class in KITTI. The new trend is to develop a refinement stage to unleash the full potential of this class of methods. The key to the refinement stage is how to effectively compute Region of Interest (ROI) features. Early works, e.g. PartA$^2$ \cite{shi2020points}, PV-RCNN \cite{shi2020pv} and VoxelRCNN \cite{deng2020voxel}, address this by first dividing ROI into a 3D grid then extracting feature at each grid location before feeding the concatenation of grid point features to a Multi-Layer Perceptron (MLP) to obtain the desired output. Their motivation is that such a grid can recover the 3D structure lost in the BEV representation used in the region proposal stage. Arguing that computing grid point features requires several hand-crafted components, CT3D \cite{sheng2021improving} devises a variant of the transformer \cite{vaswani2017attention} to compute ROI feature directly from points pooled from raw point clouds. Though having less inductive bias, CT3D achieves state-of-the-art performance, demonstrating the benefit of integrating transformer to 3D detection pipeline. 

This paper adds to the family of two-stage voxel-based 3D object detectors by making two main contributions. First, we develop a new ROI Feature Encoder (RFE) for computing per-proposal features based on Vector Attention \cite{zhao2020exploring}. RFE, together with a detection head, can serve as a refinement stage for voxel-based and point-based region proposal frameworks. Second, we observe that strong methods such as PV-RCNN \cite{shi2020pv} and CT3D \cite{sheng2021improving} employ additional modules to learn pooled point features, thus increasing model size and reducing frame rate. Therefore, we propose to pool directly from feature maps generated by the backbone during the region proposal process. Inspired by \cite{cheng2021masked}, our pooling strategy effectively fuses multi-scale features, thus increasing the model's ability to detect classes of different sizes. In addition, we conduct extensive experiments to validate the effectiveness of our method and the design choices we make. In the following, we first highlight our differences compared to closely related works in Section~\ref{sec:related-work}, then provide the conceptual details of APRO3D-Net in Section~\ref{sec:swithnet}. Section~\ref{sec:experiment} presents the implementation details and performance of our method on KITTI \cite{Geiger2012kitti} and NuScenes \cite{caesar2020nuscenes} dataset as well as conducts ablation studies. Conclusion and outlook are draw in Section~\ref{sec:conclusion}.

% In related works to compare with other transformer-based 3DOD Our use of attention mechanism in combination with a CNN backbone bring the best of both world: CNN learns localized features while attention captures long-range dependency while minimzing the negative impact of attention's quadratic complexity.

\section{related works} \label{sec:related-work}

As mentioned in Section~\ref{sec:intro}, our work belongs to the family of two-stage voxel-based detectors which comprises of PartA$^2$ \cite{shi2020points}, PV-RCNN \cite{shi2020pv}, VoxelRCNN \cite{deng2020voxel}, and CT3D \cite{sheng2021improving}. Compared to the first three methods, we are similar in the interest of using inductive bias to compensate for the loss of 3D structure in BEV representation used in the proposals generation stage. While their inductive bias is to impose a grid structure to ROI, ours takes place in position encoding of pooled points (Section~\ref{sec:pos-encoding}). Specifically, pooled points' coordinates are mapped to ROI's canonical frame and then augmented with their displacement vector to ROI's eight vertices. The difference between pooled points' augmented coordinates and that of ROI's center is input to an MLP for computing position encoding.

Our pooling strategy uses the same source as VoxelRCNN \cite{deng2020voxel} which is the intermediate feature maps generated by the 3D backbone of the region proposal framework. Instead of concatenating features pooled across different scales like VoxelRCNN, we first pool from the highest one to compute initial ROI features, then update these initial ROI features using features pooled from another feature map at a lower scale. The reason is to condition ROI features obtained at lower scales on the higher ones, thus encouraging the consistency of learned features throughout the architecture.

CT3D \cite{sheng2021improving} is the closest to our proposed approach since we share the method of computing ROI features via the attention mechanism. Compared to CT3D, we have two key differences. First, we use a different formulation of the attention mechanism, namely Vector Attention \cite{zhao2020exploring}, to assign different attention weights to different channels of one point feature. The motivation will be explained in Section~\ref{sec:attn-module}. Second, CT3D pools from raw point cloud to enable its integration into virtually any detection framework. Such flexibility comes at the cost of ignoring the valuable intermediate results of the region proposal process. This forces CT3D to recompute features for pooled points before transforming them to ROI features using the self-attention in which a pooled point feature is a weighted sum of others'. Our method pools from the backbone's intermediate feature maps. As a result, it is no longer necessary to recompute pooled features. Furthermore, by re-using backbone features, our pooling strategy maximizes the use of the information produced in the region proposal process.

It is worth noticing that there is a number of works on developing a full-transformer 3D object detectors \cite{mao2021voxel, pan20213d, misra2021end, fan2021embracing} which are orthogonal to this paper. 

\section{apro3d-Net for 3d object detection} \label{sec:swithnet}
\begin{figure*}[htb]
    \centering
    \includegraphics[width=0.65\linewidth]{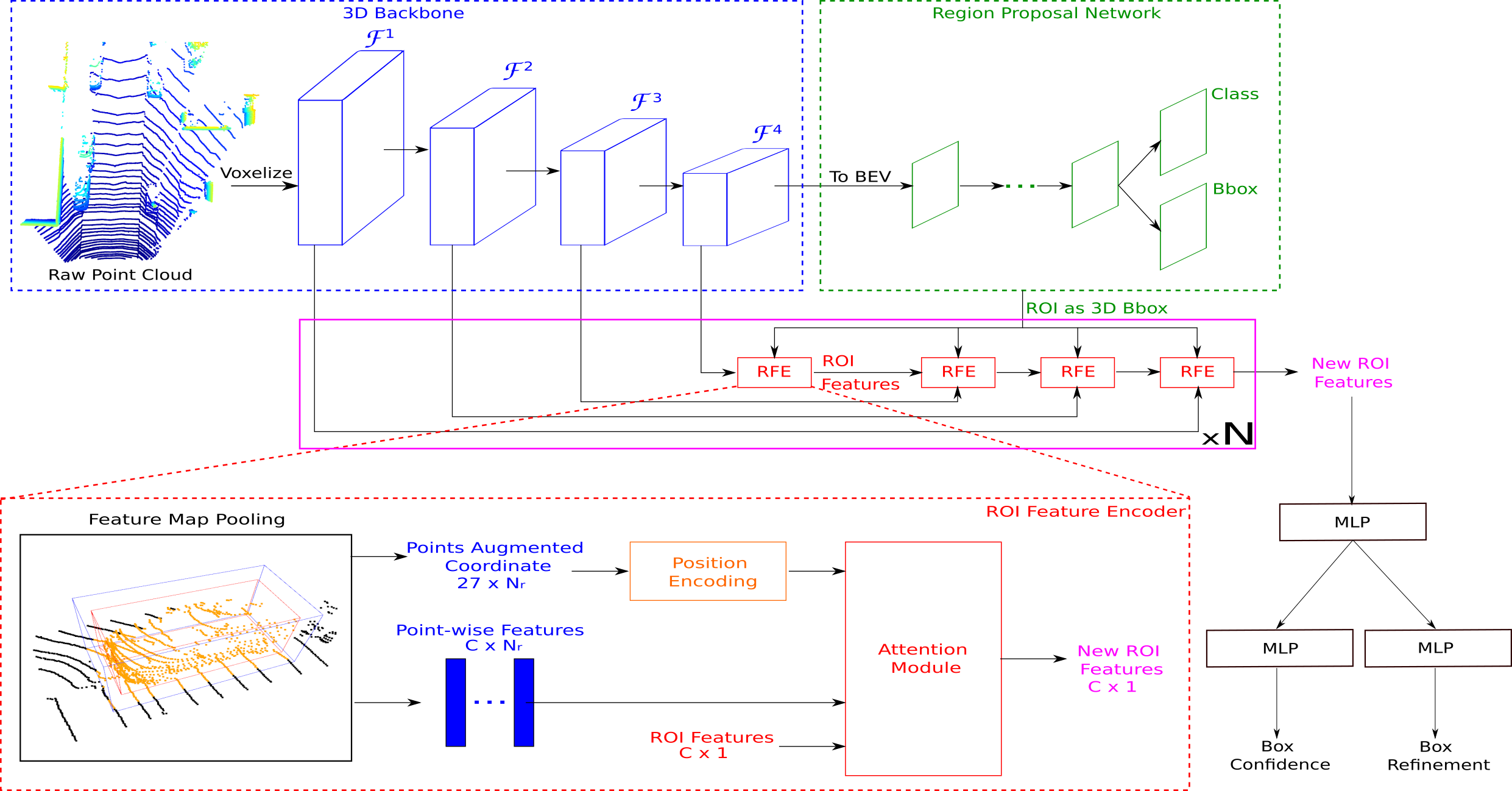}
    \caption{The overall architecture of APRO3D-Net. The voxelized point cloud is fed to a 3D backbone for feature extraction. The backbone's output is then converted to a BEV representation on which an RPN is applied to generate ROI. Several ROI Feature Encoders (RFE) transform feature maps produced by backbone into ROI features by: first pooling from inputted feature maps, then encoding pooled points position, finally refining previous ROI feature using pooled features and their position encoding via Attention Module. Refined ROI feature is mapped to confidence and refinement vector by two MLP-based detection heads. Here, blue cuboids and green parallelogram respectively denote feature maps computed by 3D and 2D convolution. Notice that the channel dimension is omitted for clarity.
    }
    \label{fig:model-overview}
\end{figure*}

%We propose a generic module, named ROI Feature Encoder (RFE), which computes ROI feature using feature maps produced during the proposal generation process. 
Figure~\ref{fig:model-overview} shows the overview of APRO3D-Net made of integrating our ROI Feature Encoder (RFE) modules to SECOND \cite{yan2018second}. Our framework first interprets feature maps created during the region proposal process into point-wise features. ROIs pool these points based on their relative locations. Pooled points features and their position encoding that incorporate an ROI's geometry are transformed into the ROI's features by the Vector Attention. The ROI's features are mapped to its confidence score and refinement vector by two MLP-made heads.

\subsection{3D Backbone and Region Proposal Network}
The reason for choosing SECOND to demonstrate our method instead of a point-based method such as PointRCNN is two folds. First, point-based methods are not as computationally efficient because of their repetitive use of query operations (ball query and k-nearest neighbor query), which can take up to $80\%$ computational time \cite{liu2019pvcnn}. More importantly, the model's final performance is strongly conditioned by how well ROIs produced by the Region Proposal Network (RPN) cover ground truth boxes. The indicator for such ability is the RPN's recall rate, which \cite{shi2020points} shows is higher with SECOND-like RPN.

SECOND first uses a backbone made of Sparse Convolutions to learn a compact representation of the input point cloud. The backbone's final output is a $C$-channel feature volume $D \times H \times W$. It is then converted into a BEV representation of size $(C \times D) \times H \times W$ by flattening the channel and depth dimension. Each location of the resulted BEV image is associated with multiple anchors corresponding to different classes and orientations. Finally, an RPN made of a standard 2D CNN predicts class probability and offset vector w.r.t associated ground truth for each anchor. Anchors modified by predicted offset vectors become Regions of Interest (ROI). ROIs are post-processed by the non-max suppression (NMS) procedure to remove redundant but low confident ROI to make the final output $\mathcal{R}$ 
\begin{equation}
    \mathcal{R} = \left\{
        \left(
        \left[
            x_r, y_r, z_r, dx_r, dy_r, dz_r, \theta_r
        \right], \text{cls}_r
        \right)
    \right\}_{r=1}^{M}
    \label{eq:roi-param}
\end{equation}
where each ROI is parameterized by the location of its center $\left[x_r, y_r, z_r\right]$, its size $\left[dx_r, dy_r, dz_r\right]$, its heading direction (i.e. yaw angle) $\theta_r$ and its class $\text{cls}_r$.   

\subsection{ROI Feature Encoder}
%RFE transforms the rich feature extracted by the backbone into ROI features. The core component of RFE is the Attention Module based on the Vector Attention \cite{zhao2020exploring} (detail will be provided later) which computing refined ROI features by cross attending to pooled features.

RFE has three sub-modules: Feature Map Pooling, Position Encoding, and Attention Module. The Feature Map Pooling interprets backbone-generated feature volumes into point-wise features and pools them according to their location relative to ROIs' bounding box. Points pooled by an ROI are assigned position encoding vectors to incorporate the ROI's geometry. The Attention Module transforms pooled points' features and their position encoding into the ROI's feature via the Vector Attention \cite{zhao2020exploring}, which essentially is a weighted sum of pooled point features.

\subsubsection{Feature Maps Pooling}
Define the LiDAR frame $\mathtt{L}$ as the following its origin is at the LiDAR's location, the X-axis coincides with the ego vehicle's heading direction, the Z-axis is the reversed gravity direction, and the Y-axis is the cross product of Z and X-axis. Without loss of generalization, assume that point clouds are expressed w.r.t this frame. An occupied voxel $\left(d, h, w\right)$ of the feature map $\mathcal{F}^i (i = 1,\hdots, 4)$ (Figure~\ref{fig:model-overview}) is interpreted into a 3D location $(x, y, z)$ by

\begin{equation}
    ^{\mathtt{L}}\begin{bmatrix}
    x, y, z
    \end{bmatrix}
    =
    \left(
    \begin{bmatrix}
    w, h, d
    \end{bmatrix} + 0.5
    \right) \cdot V^i 
    + 
    ^{\mathtt{L}}\begin{bmatrix}
    x, y, z
    \end{bmatrix}_{\text{min}}
    \label{eq:voxel2point}
\end{equation}
Here, $V^i$ is the voxel size of $\mathcal{F}^i$. It is worth noticing that $d, h, w$ are respectively the voxel's \textit{grid location} along the Z-axis, Y-axis, and X-axis. $^{\mathtt{L}}\left[x, y, z\right]_{\text{min}}$ is the minimum metric coordinate in LiDAR frame $\mathtt{L}$. Applying Eq.\eqref{eq:voxel2point} to every occupied voxel of $\mathcal{F}^i$ results in a set of point-wise features $\mathcal{P}^i = \left\{\left(^{\mathtt{L}}\mathbf{p}_j^i = \left[x_j^i, y_j^i, z_j^i\right], \mathbf{f}_j^i\right)\right\}_{j = 1}^{N^i}$. Here, $\mathbf{f}_j^i$ is the feature at the grid location gives rise to $\mathbf{p}_j^i$, while $N^i$ is the number of occupied voxels in $\mathcal{F}^i$. 

The pooling scheme, illustrated in the bottom-left corner of Figure~\ref{fig:model-overview}, is performed based on the location of point-wise features $\mathcal{P}^i$ relative to the \textit{enlarged} ROIs to incorporate missing foreground points due to the miss alignment with ground truth. Let $\mathfrak{R}_r$ denote the 3D volume occupied by an ROI $r$ after being enlarged by $\left[\Delta_x, \Delta_y, \Delta_z\right]$. A point feature $\left(\mathbf{p}_j^i, \mathbf{f}_j^i\right)$ is pooled into ROI $r$ if $\mathbf{p}_j^i \in \mathfrak{R}_r$. 

\begin{figure}[htb]
    \centering
    \includegraphics[width=0.75\linewidth]{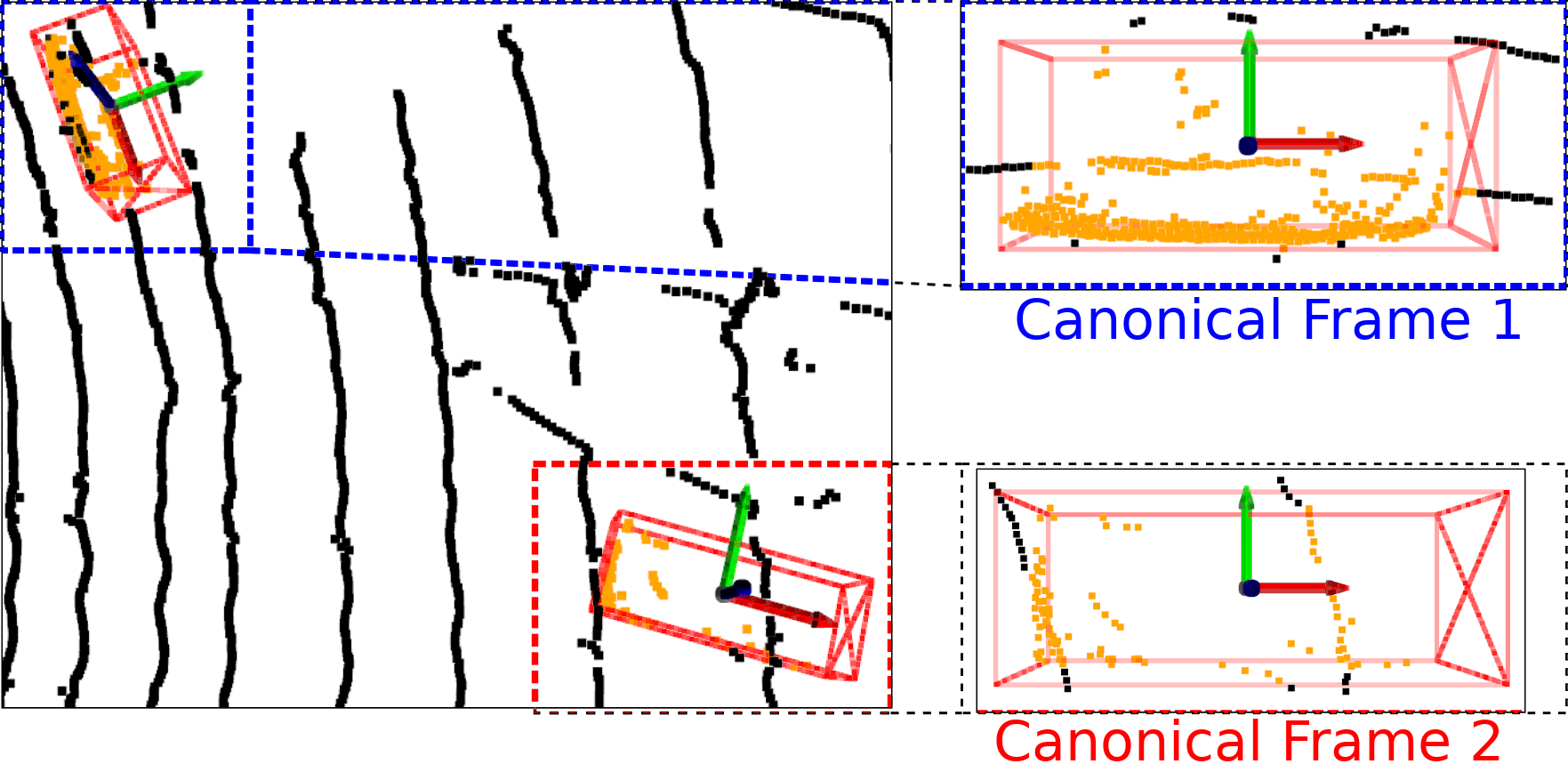}
    \caption{ROIs and pooled points in LiDAR frame (Left) compared to them in ROIs' canonical frame (Right). Here, ROI are denoted by red cuboids while pooled points are colored orange. Red, green and blue arrows respectively represents canonical frame's X, Y, Z axis.}
    \label{fig:roi-canonical}
\end{figure}

Inspired by \cite{shi2019pointrcnn, shi2020points}, we transform pooled point-wise features to ROI's canonical coordinate system to reduce the variance during training, thus improving the model's generality. This coordinate system, shown in Figure~\ref{fig:roi-canonical}, is as the following origin is at ROI's center, the X-axis has the same direction as ROI's heading direction, and the Z-axis is vertical and points upward. From Eq.\eqref{eq:roi-param}, a ROI is characterized by a seven-vector $\left[x_r, y_r, z_r, dx_r, dy_r, dz_r, \theta_r\right]$. A point $\mathbf{p}_j$ is transformed to ROI $r$'s canonical frame by

\begin{equation}
    ^{r}\mathbf{p}_j 
    =
    \begin{bmatrix}
        \cos \theta_r &  \sin \theta_r & 0 \\
        -\sin \theta_r & \cos \theta_r & 0 \\
        0 & 0 & 1
    \end{bmatrix}
    \left(
        ^{\mathtt{L}}\mathbf{p}_j^T - 
        \begin{bmatrix}
            x_r \\ y_r \\ z_r    
        \end{bmatrix}
    \right)
\end{equation}

\subsubsection{Attention Module} \label{sec:attn-module}
\paragraph{Discussion}
Once pooled, point-wise features are used to compute a single feature vector representing the entire ROI. A straightforward method is to transform points individually (via an MLP) and then synthesize them using a permutation invariance operation (e.g., sum, mean, or max pooling). However, this approach disregards valuable information about an ROI's geometry, such as points distribution or the ROI's size. To remedy this, we propose to use the attention mechanism to compute the ROI feature given pooled point-wise features. The advantage of using the attention mechanism is two folds
\begin{itemize}
    \item The model can dynamically define how much each point feature contributes to an ROI feature, thus naturally reducing the impact of background points while not suppressing them entirely. Such a balance can be helpful since background points, especially those on the ground, can provide context for estimating height.
    \item ROIs' geometry information (e.g., points location, ROI size) can be explicitly injected into the computation by position encoding.  
\end{itemize}

While the original multi-head attention \cite{vaswani2017attention} used by ViT \cite{dosovitskiy2020image} and its variants have achieved remarkable successes in the realm of computer vision, it has a drawback of treating every channel equally. In other words, a single set of scalar weights is assigned to $C$-dimension point-wise features in the weighted sum for an ROI feature. Since we pool from feature maps generated by the backbone made of convolution layers, each channel of any feature map is a detector for a certain feature \cite{zeiler2014visualizing}. Therefore, using a single set of scalar weights can risk less important features overshadowing important ones. In addition, using multi-head attention can introduce inconsistency since ViT and CNNs learn significantly different features \cite{raghu2021vision}. For the reasons above, we opt for Vector Attention \cite{zhao2020exploring} which is effective in 3D classification and segmentation tasks \cite{zhao2021point}.

\paragraph{Vector Attention}
Essentially, the computation of an ROI feature is cross-attention, where the ROI feature queries the set of pooled point-wise features. Let $\mathbf{r}$ be the initial value of the ROI feature, and $\mathcal{P}_r = \left\{\left(^{r}\mathbf{p}_j, \mathbf{f}_j\right)\right\}_{j = 1}^{N_r}$ be the set of pooled point-wise features. The new ROI feature $\hat{\mathbf{r}}$ is computed by

\begin{equation}
    \hat{\mathbf{r}} = \sum_{\mathcal{P}_r} \rho \left(
        \gamma \left(
            \varphi \left(\mathbf{r}\right) - \psi \left(\mathbf{f}_j\right) + \bm{\zeta}
        \right) 
    \right)
    \odot
    \left(
        \alpha \left(\mathbf{f}_j\right) + \bm{\zeta}
    \right)
    \label{eq:vector-attn}
\end{equation}
Here, $\varphi, \psi, \alpha$ are linear projections, $\gamma$ is an MLP, and $\rho$ is the $\mathtt{softmax}$ operation. $\odot$ denotes the Hadamard product (i.e., element-wise multiplication). $\bm{\zeta}$ represents the position encoding, whose detail will be presented shortly. In Eq.\eqref{eq:vector-attn}, $\varphi \left(\mathbf{r}\right), \psi \left(\mathbf{f}_j\right), \alpha \left(\mathbf{f}_j\right)$ respectively take the role of query, key, and value.

Using a different set of weights for each channel requires storing $N_r \times C$ parameters for computing $\hat{\mathbf{r}}$. As a result, the space complexity of the Vector Attention is $O\left(M N_r C\right)$ with $M$ is the number of ROI, thus making Vector Attention more expensive than multi-head attention. However, given that the number of ROIs in the refinement stage is relatively small thanks to NMS (100 during testing), Vector Attention is still affordable on mid-end hardware.

\begin{figure}[htb]
    \centering
    \includegraphics[width=0.95\linewidth]{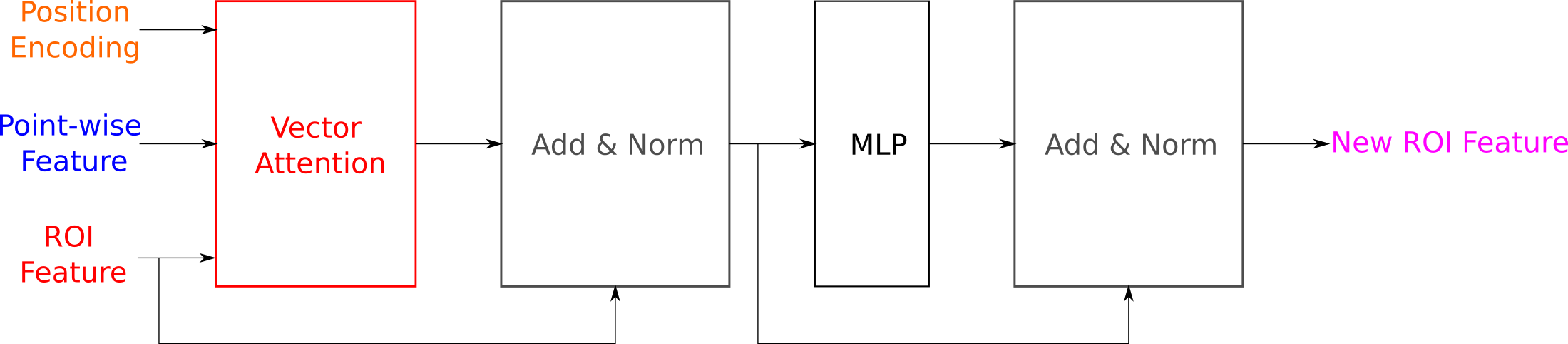}
    \caption{Architecture of the Attention Module.}
    \label{fig:attn-mod}
\end{figure}

Following \cite{vaswani2017attention}, the Attention Module, shown in Figure~\ref{fig:attn-mod}, consists of Vector Attention, residual connection, normalization layers (BatchNorm by default), and MLP.

\subsubsection{Position Encoding} \label{sec:pos-encoding}
We exploit position encoding to inject geometry information, including points location and ROI size, into the attention mechanism. A point $j$ 's location is readily available in its coordinate $^{r}\mathbf{p}_j$ in the ROI $r$'s canonical frame. To incorporate an ROI's size, we use the approach proposed by \cite{sheng2021improving} in which a point's displacement w.r.t ROI's eight vertices augments its coordinate.

\begin{equation}
    ^{r}\Tilde{\mathbf{p}}_j = 
    \begin{bmatrix}
        ^{r}\mathbf{p}_j & ^{r}\mathbf{p}_{j, 1} & \hdots & ^{r}\mathbf{p}_{j, 8}
    \end{bmatrix}
    \in \mathbb{R}^{1 \times 27}
    \label{eq:pos-embedding}
\end{equation}
In Eq.\eqref{eq:pos-embedding}, $^{r}\mathbf{p}_{j, k} (k=\left\{1, \hdots, 8\right\})$ denotes the vector going from vertex $k$ to $^{r}\mathbf{p}_j$. 

Position encoding $\bm{\zeta}$ used in Eq.\eqref{eq:vector-attn} is computed by
\begin{equation}
    \bm{\zeta} = \text{MLP}\left(
        ^{r}\Tilde{\mathbf{c}} - ^{r}\Tilde{\mathbf{p}}_j
    \right)
    \label{eq:previous-pos-enc}
\end{equation}
where $^{r}\Tilde{\mathbf{c}}$ is the result of applying Eq.\eqref{eq:pos-embedding} to the ROI's center.

\subsubsection{Handling Multi-scale Feature Maps}
While prior works pool from one fixed source such as raw point clouds \cite{sheng2021improving}, a set of sampled points \cite{shi2020pv}, or some feature maps \cite{shi2020points, deng2020voxel}, we propose to pool from every feature map. Our motivation is that each feature map has a different scale thus being helpful for detecting objects of different sizes. For example, large-scale feature maps can help detect large objects such as cars thanks to the large receptive field at each location. On the other hand, their high sparsity makes pooling with small ROIs (e.g., ROI of class pedestrians or cyclists) returns a significantly low number of points or event empty, making extracting meaningful ROI feature difficult.

Our pooling scheme is detailed by Alg.~\ref{alg:pooling-scheme}. Inspired by \cite{cheng2021masked}, we sequentially pool feature maps from the largest to the smallest scale. Once a feature map got pooled, ROI features are computed from their associated point-wise features using Eq.\eqref{eq:vector-attn}. This process repeats N times with N different sets of RFEs to increase the model's depth.

\begin{algorithm}[hbt]
    \caption{Computing ROI features by pooling from multiple feature maps} \label{alg:pooling-scheme}
    \KwIn{\\
        $\mathcal{F}^i \text{ } (i=\left\{1, \hdots, 4\right\})$ \text{: feature maps generated by backbone} \\
        $\mathcal{R} = \left\{ 
            \left( \left[ x_r, y_r, z_r, dx_r, dy_r, dz_r, \theta_r \right], \text{cls}_r  \right)
        \right\}_{r=1}^M$ \text{: set of ROI generated by RPN}
    } 
    \KwOut{
    $\mathcal{RF} = \left\{ \mathbf{r}_r \right\}$ \text{: set of ROI features} \\
    }
    \tcp{Initialize ROI features with model's learnable parameter $\Theta$}
    $\mathcal{RF} \gets \emptyset$ \;  
    \For{$r$ in $\left\{1, ..., |\mathcal{R}|\right\}$}{
        $\mathbf{r}_r \gets \Theta$ \; %\tcp*{model can have different $\Theta$ for different classes}
        $\mathcal{RF} \gets \mathcal{RF} \cup \left\{ \mathbf{r}_r \right\}$ \;
    }
    \For{$N$ times}{
        \For{$i$ in $\left\{4, ..., 1\right\}$}{
            \For{$r$ in $\left\{1, ..., |\mathcal{R}|\right\}$}{
                $\mathcal{P}_r = \left\{\left(^{r}\mathbf{p}_j, \mathbf{f}_j\right)\right\}
                \gets 
                \texttt{Pool}\left(\mathcal{F}^i\right) $ %\tcp*{pool point features from $\mathcal{F}^i$}
                
                $\mathbf{r}_r \gets \texttt{Attention}\left(\mathbf{r}_r, \mathcal{P}_r\right)$ 
                \tcp*{Eq.\eqref{eq:vector-attn}}
            }
        }
    }
\end{algorithm}

\subsection{Detection Heads and Learning Targets}
An ROI's feature computed by a series of RFEs is mapped to a higher dimension space by a two-hidden layer MLP before decoded into ROIs' confidence and refinement vector. Following \cite{shi2020points}, an ROI's confidence is set to the normalized Intersection over Union (IoU) with its associated ground truth, thus making the refinement stage class-agnostic. Let $\mathtt{IoU}$ denote ROI $r$'s regular IoU, its normalized IoU is 
\begin{equation}
    c_r^{*} = \begin{cases}
        1 & \text{ if } \mathtt{IoU} > \chi_H \\
        0 & \text{ if } \mathtt{IoU} < \chi_L \\
        \frac{\mathtt{IoU} - \chi_L}{\chi_H - \chi_L} & \text{ otherwise}
    \end{cases}
    \label{eq:norm-iou}
\end{equation}
where, $\chi_H$ and $\chi_L$ are foreground and background threshold.

The target of the refinement head is the normalized residue of an ROI w.r.t its associated ground truth. Given the parameters of ROI $r$ defined by Eq.\eqref{eq:roi-param} and its associated ground truth, its normalized residue $\bm{\delta_r}^* = \left[x^*, y^*, z^*, dx^*, dy^*, dz^*, \theta^*\right]$ is  
\begin{equation}
    \begin{matrix}
    x^* = \frac{x_r^g - x_r}{d}, & y^* = \frac{y_r^g - y_r}{d}, & z^* = \frac{z_r^g - z_r}{dz_r} \\
    dx^* = \log \frac{dx_r^g}{dx_r}, & dy^* = \log \frac{dy_r^g}{dy_r}, & dz^* = \log \frac{dz_r^g}{dz_r} \\
    \theta^* = \theta_r^g - \theta_r
    \end{matrix}
    \label{eq:target-refine}
\end{equation}
In Eq.\eqref{eq:target-refine}, the superscript $g$ denotes the ground truth box's parameters, while the subscript $r$ represents ROI index. $d = \sqrt{x_r^2 + y_r^2}$ is the diagonal of the base of the ROI. 

\subsection{Loss Function}
Our RFE module can be trained end-to-end with the RPN by optimizing the summation of the RPN loss, the refinement stage's loss, and an auxiliary loss.

\begin{equation}
    \mathcal{L} = \mathcal{L}_{\text{RPN}} + \mathcal{L}_{\text{refine}} + \mathcal{L}_{\text{aux}}
\end{equation}

\paragraph{RPN Loss} 
Since we adopt the backbone and the RPN of SECOND, the RPN loss is as in \cite{yan2018second} which is the sum of classification and a regression loss
\begin{equation}
    \mathcal{L}_{\text{RPN}} = \frac{1}{A_{+}} \sum_{a} \left[
        \mathcal{L}_{\text{cls}} \left(c_a, c_a^*\right)
        +
        \mathds{1}\left(c_a^* \neq 0\right) \mathcal{L}_{\text{reg}} \left(\delta_a, \delta_a^*\right)
    \right]
    \label{eq:loss-rpn}
\end{equation}
where, $c_a$ and $\delta_a$ are output of RPN's Class branch and Bbox branch, $A_{+}$ is the number of positive anchors. The classification target of the RPN $c_a^*$ is the class of ground truth box of anchor $a$. The regression target $\delta_a^*$ of anchor $a$ is calculated according to Eq.\eqref{eq:target-refine}. $\mathds{1}\left(c_a^* \neq 0\right)$ is the indicator function which takes value of $1$ for positive anchors whose $c_a^* \neq 0$  and $0$ otherwise. The classification loss $\mathcal{L}_{\text{cls}}$ is the Focal Loss \cite{lin2017focal}, while the regression loss $\mathcal{L}_{\text{reg}}$ is the Huber Loss (i.e. smooth-L1).

\paragraph{Refining Loss}
Similar to the RPN loss, this loss makes of classification and a regression loss.
\begin{equation}
    \mathcal{L}_{\text{refine}} = \frac{1}{M} \sum_{r} \left[
        \mathcal{L}_{\text{cls}} \left(c_r, c_r^*\right)
        +
        \mathds{1}\left(c_r^* \geq \chi_{reg}\right) \mathcal{L}_{\text{reg}} \left(\delta_r, \delta_r^*\right)
    \right]
    \label{eq:loss-refining}
\end{equation}
Here, the classification target is the normalized IoU (Eq.\eqref{eq:norm-iou}) and $\mathcal{L}_{\text{refine}}$ is normalized by the total number of ROIs $M$. The regression threshold $\chi_{reg}$ used in Eq.\eqref{eq:loss-refining} is different to the foreground threshold $\chi_H$ of Eq.\eqref{eq:norm-iou}.

\paragraph{Auxiliary Loss}
Inspired by \cite{he2020structure, shi2020points}, an auxiliary supervision is applied to two backbone-generated feature maps, $\mathcal{F}^3$ and $\mathcal{F}^4$, to guide the feature extraction. To be specific, they are interpreted into point-wise features. Each point feature $k$ is then fed into an MLP to predict its foreground probability $f_k$, offset toward associated ground truth box's center $o_k$, and part probability $p_k$ \cite{shi2020points}. A point is labeled as foreground if it is inside a ground truth box. The label of the part probability of foreground points is essentially their coordinate in the canonical frame (Figure~\ref{fig:roi-canonical}) of the associated ground truth box normalized by the box's sizes.

\begin{equation}
\begin{split}
    \mathcal{L}_{\text{aux}} &= 
    \frac{1}{P_+} \left[
        \sum_{k=1}^P\mathcal{L}_{\text{cls}}\left(f_k, f_k^*\right)
    \right] + \\
    & \frac{1}{P_+} \left[
        \sum_{k=1}^{P_+} \mathcal{L}_{\text{reg}}\left(o_k, o_k^*\right)
        + \mathcal{L}_{\text{bce}}\left(p_k, p_k^*\right)
    \right]
\end{split}
\label{eq:loss-aux}
\end{equation}
In Eq.\eqref{eq:loss-aux}, $\mathcal{L}_{\text{bce}}$ is the Binary Cross Entropy (BCE) loss. The $*$ in the superscript denotes the label, while subscript $k$ is the point index. $P_+$ is the number of foreground points. The regression loss and BCE loss are only calculated for foreground points.

\section{Experiments} \label{sec:experiment}
To demonstrate the effectiveness of our method, we evaluate it on KITTI \cite{Geiger2012kitti} and NuScenes \cite{caesar2020nuscenes} datasets. Furthermore, we carry out comprehensive ablation studies to understand the influence of each module on the overall performance.

\subsection{Datasets}
The KITTI Dataset contains 7481 and 7581 samples for training and testing. Each sample comprises sensory measurements collected by a LiDAR and several cameras. A common practice when working with KITTI is to split the original training data into 3712 training samples and 3769 validation samples for experimental studies. On the other hand, we adjust the training to validation ratio to 4:1 when preparing the submission to the official test server for benchmarking.

Compared to KITTI, NuScenes is more challenging due to its larger size and requirement of detecting more classes. Specifically, NuScenes offers 28130 training and 6019 validation samples. Each sample, or keyframe, comprises data collected by one LiDAR, six cameras, and five radars when they are in sync. Annotations are provided for 23 object classes, among which 10s are chosen for the detection task.

\subsection{Implementation Details}
We build our method to work on top of SECOND (or other 3D proposals methods). For efficiency, we use the implementation of SECOND and other RPNs provided by the OpenPCDet \cite{openpcdet2020} toolbox. To demonstrate the robustness of our choice of model's hyperparameters, we keep them constant for experiments on both KITTI and NuScenes. The three exceptions are the point cloud range, the initial voxel size and the number of channels of the last feature map, $\mathcal{F}^4$.

\subsubsection{RPN}
Since we do not introduce any modification to SECOND, the following presents the parameters directly related to our method, while the rest can be found in OpenPCDet. 

\begin{table*}[htb]
\centering
\caption{Performance comparison on KITTI \textit{test} set with AP calculated with 40 recall positions}
\begin{tabular}{c | c ||c c c | c c c} 
    \hline
    \multirow{2}{4em}{Method} & Num Para- &  \multicolumn{3}{c|}{Car - 3D Detection} &  \multicolumn{3}{c}{Cyclist - 3D Detection} \\
    & meters (M) & Easy & Mod. & Hard  & Easy & Mod. & Hard \\ 
    \hline
    SECOND \cite{yan2018second} & 20 & 83.34 & 72.55 & 65.82 & 71.33 & 52.08 & 45.83 \\
    PointPillar \cite{lang2019pointpillars} & 18 & 82.58 & 74.31 & 68.99 & 77.10 & 58.65 & 51.92 \\
    PointRCNN \cite{shi2019pointrcnn} & 16 & 86.96 & 75.64 & 70.70 & 74.96 & 58.82 & 52.53 \\
    SA-SSD \cite{he2020structure} & 226 & 88.75 & 79.79 & 74.16 & - & - & -\\
    Part $A^2$ \cite{shi2020points} & 40.8 & 87.81 & 78.49 & 73.51 & - & - & -\\
    PV-RCNN \cite{shi2020pv} & 50 & 90.25 & 81.43 & 76.82 & \textbf{78.60} & 63.71 & 57.65 \\
    Voxel R-CNN \cite{deng2020voxel} & 28 & \textbf{90.90} & 81.62 & 77.06 & - & - & - \\
    CT3D \cite{sheng2021improving} & 30 & 87.83 & \textbf{81.77} & \textbf{77.16} & - & - & -\\
    \hline
    APRO3D-Net (ours) & 22.4 & 87.09 & 80.30 & 76.10 & 78.54 & \textbf{64.55} & \textbf{57.78} \\
    \hline
\end{tabular}
\label{perf:kitti-test}
\end{table*}

\paragraph{KITTI} Point clouds are clipped by $[0\text{m}, 70.4\text{m}]$ in the X-axis, $[-40\text{m}, 40\text{m}]$ in the Y-axis, and  $[-3\text{m}, 1\text{m}]$ in the Z-axis and voxelized with grid size of $[0.05\text{m}, 0.05\text{m}, 0.1\text{m}]$. SECOND's intermediate feature maps ($\mathcal{F}^i, i=1,\hdots,4$) have 16, 32, 64, and 64 channels. Proposals are post-processed by NMS with the overlapped threshold of 0.8 (0.7) to obtain 512 (100) ROIs during training (testing).

\paragraph{NuScenes} point clouds' range $[-51.2\text{m}, 51.2\text{m}]$ along X-axis, Y-axis, and $[-5\text{m}, 3\text{m}]$ along Z-axis. The voxel size for discretizing the input point cloud is $[0.1\text{m}, 0.1\text{m}, 0.2\text{m}]$. The point cloud of a NuScenes keyframe (i.e., sample) contains about 40k points which are just one-third the size of a KITTI point cloud, thus making it highly difficult for any methods. We follow the common practice that maps point cloud in 10 previous non-keyframe to the timestamp of the keyframe using ego vehicle's odometry to increase the number of points by ten times. Regarding the 3D backbone, the number of channels in the last feature map $\mathcal{F}^4$ is 128, while the rest are similar to the KITTI configuration.

\subsubsection{ROI Feature Encoder}
128 ROIs are sampled from RPN output for the refinement stage during training. Each ROI is then enlarged by $0.5m$ along three dimensions for pooling.  We empirically find that the pooling from second feature map, $\mathcal{F}^2$, does not bring significant improvement to the final performance. Therefore, we opt for pooling 64, 128, and 256 points from $\mathcal{F}^4, \mathcal{F}^3, \mathcal{F}^1$ for each ROI. 

The feature dimension is kept constant at $d_a$ equal to 128 throughout the Attention Module. Features pooled from $\left\{\mathcal{F}^4, \mathcal{F}^3, \mathcal{F}^1\right\}$ are linearly mapped to $d_a$ before going to the Vector Attention. The MLP of the Attention Module and Position Encoding has a 256-neuron hidden layer activated ReLU function. The sequential pooling from $\mathcal{F}^4$ to $\mathcal{F}^1$ for computing ROIs' feature is repeated 3 times with three different sets of RFEs.

\subsubsection{Training}
The entire architecture presented in Figure~\ref{fig:model-overview} is optimized end-to-end by the Adam optimizer. For KITTI, we train the model for 100 epochs with a total batch size of 24. The learning rate is controlled by One-Cycle policy \cite{smith2019super} with a maximum value of $0.01$. For NuScenes, the training lasts 20 epochs while the same batch size and the learning rate policy remain unchanged. The maximum learning rate reduces to $0.03$. In the detection head, the foreground threshold $\chi_H$, background threshold $\chi_L$, and regression IoU threshold $\chi_{\text{reg}}$ are $0.75, 0.25$ and $0.55$. We use the same data augmentation strategy as \cite{yan2018second, shi2020pv, shi2020points}. 

\subsection{Results On KITTI Dataset}
The detection task of KITTI dataset concerns 3 classes: Car, Pedestrian and Cyclist. performance is measured by the Average Precision (AP) metric computed at 40 recall positions \footnotemark  $\text{ }$ with an IoU threshold of 0.7 for Car and 0.5 for others. The comparison of our method against the state-of-the-art on KITTI \textit{test} set is presented in Table~\ref{perf:kitti-test}. Among methods built on top SECOND, namely \cite{he2020structure, shi2020pv, deng2020voxel, sheng2021improving}, we achieve the best performance in class Cyclist and competitive performance in class Car, passing the 80 AP threshold, while having the least number of parameters. Note that we train a single model for two classes instead of separate models as previously done by \cite{yan2018second, lang2019pointpillars, shi2019pointrcnn}.

\footnotetext[2]{Since 08.10.2019, the number of recall position for computing AP has been increased from 11 to 40.}

\footnotetext[3]{Performance of model trained for 3 classes, reproduced from official code release}

Our performance on KITTI \textit{val} set with AP calculated at 40 recall positions is also reported in Table~\ref{perf:kitti-val-40}, indicating that the gap between our method and top performers in class Car is shortened, with the highest difference being just 0.45 AP. Compared to CT3D which also computes ROIs' feature using the attention mechanism, we surpass their AP for class Pedestrian and Cyclist by $1.42$ and $1.47$. %We also presents our performance calculated at 11 recall position in Table\ref{perf:kitti-val-11} for comparison against methods which don't report theirs on the 40 recall setting.

\begin{table}[htb]
\centering
\caption{Performance comparison on KITTI \textit{val} set with AP calculated at 40 recall positions}
\begin{tabular}{c||c c c}
    \hline
    \multirow{2}{4em}{Method} &  \multicolumn{3}{c}{$\text{AP}_{\text{3D}}$ - Moderate} \\
    & Car & Cyclist & Pedestrian \\ 
    \hline
   % SA-SSD \cite{he2020structure} & 84.54 & - & - \\
    PV-RCNN \cite{liu2019pvcnn} & 84.83 & 71.95 & 56.67 \\
    Voxel R-CNN \cite{deng2020voxel} & \textbf{85.29} & - & - \\
    Voxel R-CNN \footnotemark & 84.95 & 71.43 & \textbf{58.24} \\
    CT3D \cite{sheng2021improving} & 84.97 & 71.88 & 55.58 \\
    \hline
    APRO3D-Net (ours) & 84.85 & \textbf{73.35} & 57.41 \\
    \hline
\end{tabular}
\label{perf:kitti-val-40}
\end{table}

% \begin{table}[htb]
% \centering
% \caption{Performance comparison on KITTI \textit{val} set with AP calculated at 11 recall positions}
% \begin{tabular}{c||c c c}
%     \hline
%     \multirow{2}{4em}{Method} &  \multicolumn{3}{c}{$\text{AP}_{\text{3D}}$ - Moderate} \\
%     & Car & Pedestrian & Cyclist \\ 
%     \hline
%     SECOND \cite{yan2018second} & 78.62 & 52.98 & 67.15 \\
%     PointPillar \cite{lang2019pointpillars} & 77.28 & 52.29 & 62.68 \\
%     PointRCNN \cite{shi2019pointrcnn} & 78.70 & 54.41 & 72.11 \\
%     SA-SSD \cite{he2020structure} & 79.41 & 48.01 & 63.37 \\
%     Part $A^2$ \cite{shi2020points} & 79.40 & \textbf{60.05} & 69.90 \\
%     PV-RCNN \cite{liu2019pvcnn} & 83.61 & 57.90 & 70.47 \\
%     Voxel R-CNN \cite{deng2020voxel} & 84.54 & - & - \\
%     CT3D \cite{sheng2021improving} & \textbf{85.04} & 56.28 & 71.71 \\
%     \hline
%     Ours & 83.51 & 57.45 & \textbf{72.97} \\
%     \hline
% \end{tabular}
% \label{perf:kitti-val-11}
% \end{table}

\begin{table*}[htb]
    \centering
    \caption{AP on NuScenes dataset}
    \begin{tabular}{c||c|c|c|c|c|c|c|c|c|c|c }
    \hline
    Method & Car & Ped & Bus & Barrier & Traf. Cone & Truck & Trailer & Motor & Cons. Veh. & Bicycle & mAP  \\
    \hline
    SECOND \cite{yan2018second}
    & 75.53 & 59.86 & 29.04 & 32.21 & 22.49 & 21.88 & 12.96 & 16.89 & 0.36 & 0 & 27.12 \\
    %\hline
    PointPillars \cite{lang2019pointpillars}
    & 70.5 & 59.9 & 34.4 & 33.2 & 29.6 & 25.0 & 20.0 & 16.7 & 4.5 & 1.6 & 29.5 \\
    %\hline
    3DSSD \cite{yang20203dssd}
    & \textbf{81.20} & 70.17 & 61.41 & 47.94 & 31.06 & \textbf{47.15} & 30.45 & 35.96 & 12.64 & 8.63 & 42.66 \\
    %\hline
    InfoFocus \cite{wang2020infofocus}
    & 77.6 & 61.7 & 50.5 & 43.4 & 33.4 & 35.4 & 25.6 & 25.2 & 8.3 & 2.5 & 36.4 \\
    \hline
    APRO3D-Net (ours)
    & 77.75 & \textbf{74.02} & \textbf{64.86} & \textbf{52.61} & \textbf{46.34} & 43.99 & \textbf{34.9} & \textbf{39.36} & \textbf{13.44} & \textbf{23.00} & \textbf{47.03} \\
    \hline
    \end{tabular}
    \label{perf:nuscenes-val}
\end{table*}

\subsection{Results On NuScenes Dataset}
The main metric used by NuScenes is the Average Precision (AP) score, which is computed as the normalized area under the precision-recall curve with a minimum recall of 0.1. Instead of using IoU as matching criteria like KITTI, NuScenes use the euclidean distance between the center of a predicted box and its ground truth. The performance on NuScenes \textit{validation} set is shown in Table~\ref{perf:nuscenes-val}.  In this table, SECOND and PointPillars are single-stage methods, while others, including ours, are two-stage. In this more challenging method, the benefit of integrating our RFE to SECOND is more prominent, indicated by almost 20 mAP improvements. In addition, ours outperforms 3DSSD \cite{yang20203dssd} and InfoFocus \cite{wang2020infofocus}, two recent two-stage methods, by a large margin except for class Car and Truck. This competitive performance on NuScenes dataset shows our method's ability to handle object classes with high variance in scales.

\subsection{Qualitative Performance}
\begin{figure}[htb]
    \centering
    \includegraphics[width=0.95\linewidth]{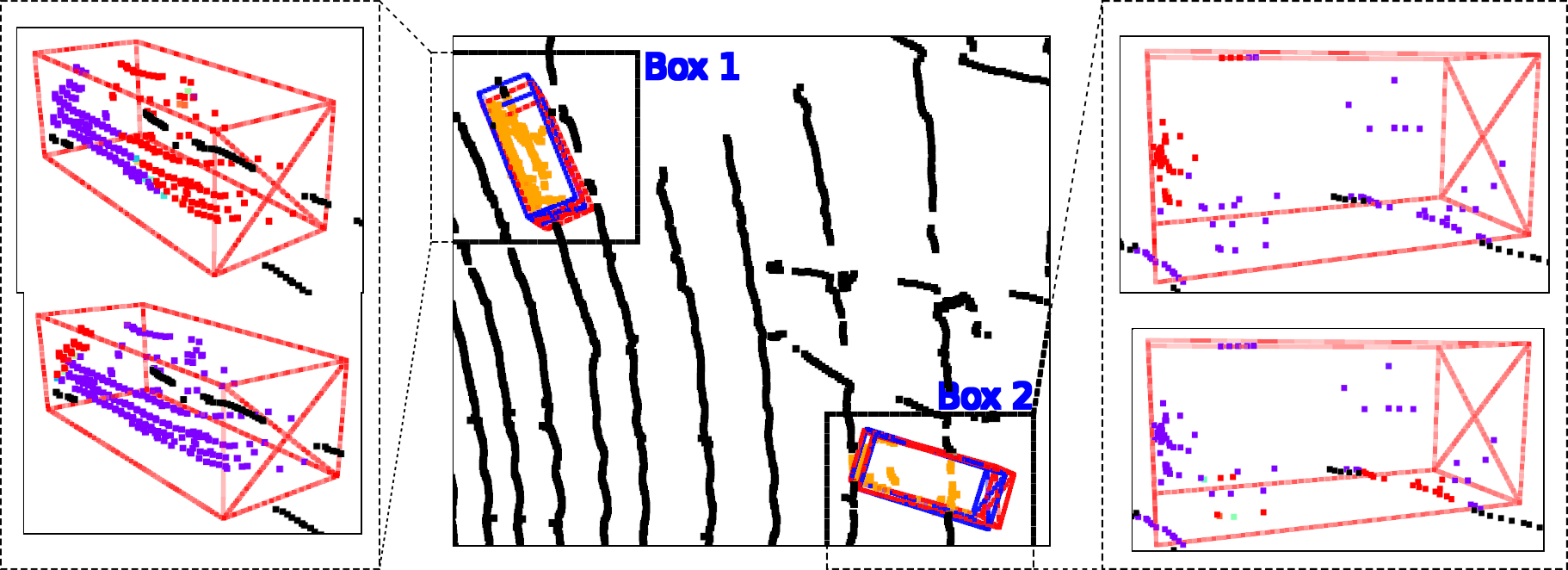}
    \caption{Visualization of attention weights. Predictions and their associated ground truth boxes are respectively marked by red and blue. Orange denotes pooled points. In two zoomed window, points are color coded according to their attention weights. The hotter the color, the higher attention weight.}
    \label{fig:vis-roi-attn-weights}
\end{figure}

\begin{figure*}[htb]
    \centering
    \includegraphics[width=0.8\linewidth]{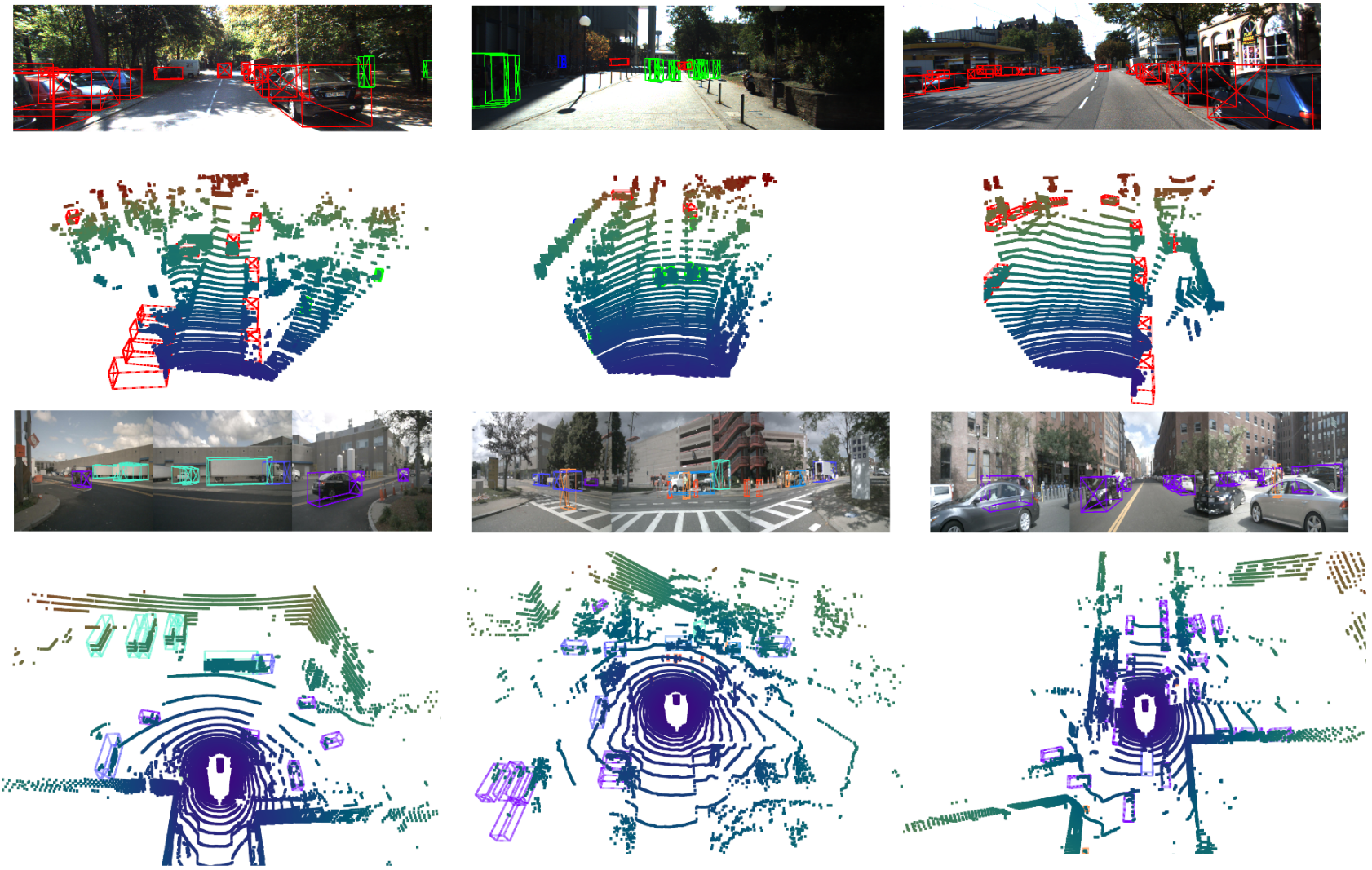}
    \caption{Visualization of prediction made by APRO3D-Net on the \textit{test} split of KITTI (upper row) and NuScenes (lower row) dataset.}
    \label{fig:kitti-test-qualitative}
\end{figure*}

To show that different channels within the same point feature contribute differently to an ROI feature, we visualize pooled points' attention weight (the term on the left of $\odot$ in Eq.\eqref{eq:vector-attn}) in Figure~\ref{fig:vis-roi-attn-weights}. As can be seen, the region of ROI, where attention is concentrated, varies across channels. For example, Box 1 respectively pays most attention to its front and rear to compute two different channels of its feature. In addition, a visual evaluation of our method's performance on the \textit{test} split of KITTI and NuScenes dataset made by projecting predictions onto images, as in Figure~\ref{fig:kitti-test-qualitative}, shows good result.

\subsection{Ablation studies}
To validate our design choices and understand the impact of each module on the overall performance, we perform extensive ablation studies. All models used in this section are trained on KITTI \textit{train} set and evaluated on KITTI \textit{val} set. Unless stated otherwise, evaluations are based on AP calculated at 40 recall precision.

First, we verify our motivation for choosing Vector Attention over multi-head attention by changing the attention formula in Eq.\eqref{eq:vector-attn} while keeping the rest of the architecture unchanged. The result shown in Table~\ref{abl:multi-head-vs-vector} confirms the superiority of Vector Attention with significant AP difference for class Car and Cyclist. Such difference is due to Vector Attention enabling the model to choose where to look (which points) and what to look for (which channels) when computing ROI features, as illustrated in Figure~\ref{fig:vis-roi-attn-weights}. On the other hand, multi-head attention can only choose where to look because of its scalar weight.

\begin{table}[htb]
\centering
\caption{performance on kitti \textit{val} set of multi-head attention compared to vector attention}
\begin{tabular}{c||c c c }
    \hline
    \multirow{2}{4em}{Method} &  \multicolumn{3}{c}{$\text{AP}_{\text{3D}}$ - Moderate} \\
    & Car & Cyclist & Pedestrian \\ 
    \hline
    Multi-head Attention & 82.50 & 70.35 & 57.58 \\
    Vector Attention & 84.85 & 73.35 & 57.41 \\
    \hline
    Improvement & \textbf{2.35} & \textbf{3.00} & -0.17 \\
    \hline
\end{tabular}
\label{abl:multi-head-vs-vector}
\end{table}

The second experiment is to analyzes the impact of position encoding on the overall performance. The first row of Table~\ref{abl:pos-enc} shows the result of the model that does not use position encoding, meaning setting $\bm{\zeta}$ in Eq.\eqref{eq:vector-attn} to $0$. The second row is the performance of building position encoding from the point displacement relative to ROI center only. In other words, $^{r}\mathbf{p}_{j, 1}, \hdots, ^{r}\mathbf{p}_{j, 8}$ are removed from Eq.\eqref{eq:pos-embedding}. Table~\ref{abl:pos-enc} shows that using position encoding increase performance for class Car, Cyclist, and Pedestrian to 8.19, 6.24, 4.03 AP. Moreover, position encoding contains the most information about ROI geometry (third row) performs the best overall, especially for the most important class Car.

\begin{table}[htb]
\centering
\caption{performance on kitti \textit{val} set of different position encoding methods}
\begin{tabular}{c||c c c }
    \hline
    \multirow{2}{4em}{Method} &  \multicolumn{3}{c}{$\text{AP}_{\text{3D}}$ - Moderate} \\
    & Car & Cyclist & Pedestrian \\ 
    \hline
    None & 76.66 & 69.65  & 53.38 \\
    
    Center & 82.72 & \textbf{75.89} & 55.74 \\
    
    Center and Vertices & \textbf{84.85} & 73.35 & \textbf{57.41} \\
    \hline
\end{tabular}
\label{abl:pos-enc}
\end{table}

Next, three pooling strategies are compared. The performance shown in the first row of Table~\ref{abl:pooling} is obtained by equally pooling $M$ points from each feature map $\mathcal{F}^i$ then concatenating their features before passing them to RFEs for ROIs feature computation. In the other rows, we sequentially pool from $\mathcal{F}^4$ to $\mathcal{F}^1$ while skipping $\mathcal{F}^2$. The difference between second and third row is the sequential pooling process takes place only once in the second row while it repeats three times in the third. In other words, $N$ of Alg.~\ref{alg:pooling-scheme} is set to 1 and 3 in row second and third, respectively. Even though pooling all at once (first row) does not show any significant performance drop for class Car while achieving the best performance in class Pedestrian, this pooling strategy is the most memory intensive. The number of points to be processed, which linearly grows with the number of feature maps and ROIs, can quickly overflow GPUs' memory. Another aspect to be noticed is repeating the pooling process (with a different set of RFE) only brings marginal performance gain.

\begin{table}[htb]
\centering
\caption{performance on kitti \textit{val} set of different pooling methods}
\begin{tabular}{c||c c c }
    \hline
    \multirow{2}{4em}{Method} &  \multicolumn{3}{c}{$\text{AP}_{\text{3D}}$ - Moderate} \\
    & Car & Cyclist & Pedestrian\\ 
    \hline
    All at once & 84.15 & 71.04 & \textbf{57.55}\\
    
    Sequential without repetition & 84.66 & \textbf{75.34} & 56.15 \\
    
    Sequential with repetition & \textbf{84.85} & 73.35 & 57.41 \\ % allow for pooling more points at each
    % stage
    \hline
\end{tabular}
\label{abl:pooling}
\end{table}

Finally, the versatility of our method is demonstrated by its integration with different RPNs: SECOND, PartA$^2$, PointRCNN. Since PartA$^2$ has a UNet-like backbone, we pool from feature maps produced by its up-sampling branch while keeping the rest of the architecture unchanged. In the case of PointRCNN, we pool from the final output of its backbone. Note that when not using the RFE, PointRCNN and PartA$^2$ use their own refinement stage. Table~\ref{abl:backbone} shows the performance gain at different difficulty levels of class Car, thus confirming the effectiveness of our method. The limited gain when integrating with PointRCNN can be explained by its lower recall rate compared to SECOND. This experiment validates our design choice regarding the RPN method.

\begin{table}[htb]
\centering
\caption{gain of performance on kitti \textit{val} set brought by rfe to different rpns}
\begin{tabular}{c||c c c }
    \hline
    \multirow{2}{5em}{Method} &  \multicolumn{3}{c}{$\text{AP}_{\text{3D}}$ (Car)} \\
    & Easy & Moderate & Hard\\ 
    \hline
    SECOND & 88.61 & 78.62 & 77.22 \\
    SECOND + RFE & \textbf{+0.75} & \textbf{+4.89} & \textbf{+1.56}  \\
    \hline
    
    PartA$^2$ & 89.47 & 79.47 & 78.54 \\
    PartA$^2$ + RFE & -0.08 & \textbf{+3.16} & \textbf{+0.36} \\
    \hline
    
    PointRCNN & 88.88 & 78.63 & 77.38 \\ 
    PointRCNN + RFE& \textbf{+0.06} & \textbf{+0.38} & \textbf{+1.02} \\
    \hline
\end{tabular}
\label{abl:backbone}
\end{table}

\section{Conclusion} \label{sec:conclusion}
In conclusion, this paper develops a proposals refinement stage for 3D object detection. The core of this stage is the RFE module which transforms pooled points features into ROI features using Vector Attention. In addition, we propose a pooling strategy that effectively fuses multi-scale features extracted by the 3D backbone, thus increasing the model's ability to handle objects of different sizes. Experiments on KITTI and NuScenes datasets validate the effectiveness of our method. 

As for future work, we would like to extend the RFE module to enable the fusion of LiDAR with other sensing modalities such as cameras or radars. Another possibility is to explore how ROI features extracted by RFEs can be used for tracking.  
%Our preliminary idea is to employ a contrastive loss to make features of ROIs corresponding to the same physical object be as similar as possible while being significantly different to ROI features of different objects. If ROI features are sufficiently distinctive, the multi-object tracking task can be addressed as a bipartite matching problem with the cost function being the difference between a pair of ROI features.

%%%%%%%%%%%%%%%%%%%%%%%%%%%%%%%%%%%%%%%%%%%%%%%%%%%%%%%%%%%%%%%%

\section*{ACKNOWLEDGMENT}
This work was carried out in the framework of the NExT Senior Talent Chair DeepCoSLAM, which were funded by the French Government, through the program Investments for the Future managed by the National Agency for Research (ANR-16-IDEX-0007), and with the support of Région Pays de la Loire and Nantes Métropole. This work was granted access to the HPC resources of IDRIS under the allocation 2021-AD011012128R1 made by GENCI. This work has been supported in part by the ANR AIby4 under the number ANR-20-THIA-0011.

%%%%%%%%%%%%%%%%%%%%%%%%%%%%%%%%%%%%%%%%%%%%%%%%%%%%%%%%%%%%%%%%%%%%%%%%%%%%%%%%

%\bibliographystyle{IEEEtran}
%\bibliography{IEEEabrv, ref}

\begin{thebibliography}{10}
\providecommand{\url}[1]{#1}
\csname url@rmstyle\endcsname
\providecommand{\newblock}{\relax}
\providecommand{\bibinfo}[2]{#2}
\providecommand\BIBentrySTDinterwordspacing{\spaceskip=0pt\relax}
\providecommand\BIBentryALTinterwordstretchfactor{4}
\providecommand\BIBentryALTinterwordspacing{\spaceskip=\fontdimen2\font plus
\BIBentryALTinterwordstretchfactor\fontdimen3\font minus
  \fontdimen4\font\relax}
\providecommand\BIBforeignlanguage[2]{{%
\expandafter\ifx\csname l@#1\endcsname\relax
\typeout{** WARNING: IEEEtran.bst: No hyphenation pattern has been}%
\typeout{** loaded for the language `#1'. Using the pattern for}%
\typeout{** the default language instead.}%
\else
\language=\csname l@#1\endcsname
\fi
#2}}

\bibitem{chen2017mv3d}
X.~Chen, H.~Ma, J.~Wan, B.~Li, and T.~Xia, ``Multi-view 3d object detection
  network for autonomous driving,'' in \emph{Proceedings of the IEEE/CVF
  Conference on Computer Vision and Pattern Recognition}, 2017, pp. 1907--1915.

\bibitem{yang2018pixor}
B.~Yang, W.~Luo, and R.~Urtasun, ``Pixor: Real-time 3d object detection from
  point clouds,'' in \emph{Proceedings of the IEEE conference on Computer
  Vision and Pattern Recognition}, 2018, pp. 7652--7660.

\bibitem{zhou2018voxelnet}
Y.~Zhou and O.~Tuzel, ``Voxelnet: End-to-end learning for point cloud based 3d
  object detection,'' in \emph{Proceedings of the IEEE conference on computer
  vision and pattern recognition}, 2018, pp. 4490--4499.

\bibitem{yan2018second}
Y.~Yan, Y.~Mao, and B.~Li, ``Second: Sparsely embedded convolutional
  detection,'' \emph{Sensors}, vol.~18, no.~10, p. 3337, 2018.

\bibitem{lang2019pointpillars}
A.~H. Lang, S.~Vora, H.~Caesar, L.~Zhou, J.~Yang, and O.~Beijbom,
  ``Pointpillars: Fast encoders for object detection from point clouds,'' in
  \emph{Proceedings of the IEEE/CVF Conference on Computer Vision and Pattern
  Recognition}, 2019, pp. 12\,697--12\,705.

\bibitem{shi2019pointrcnn}
S.~Shi, X.~Wang, and H.~Li, ``Pointrcnn: 3d object proposal generation and
  detection from point cloud,'' in \emph{IEEE/CVF Conference on Computer Vision
  and Pattern Recognition (CVPR),}, 2019, pp. 770--779.

\bibitem{yang2019std}
Z.~Yang, Y.~Sun, S.~Liu, X.~Shen, and J.~Jia, ``Std: Sparse-to-dense 3d object
  detector for point cloud,'' in \emph{Proceedings of the IEEE/CVF
  International Conference on Computer Vision}, 2019, pp. 1951--1960.

\bibitem{yang20203dssd}
Z.~Yang, Y.~Sun, S.~Liu, and J.~Jia, ``3dssd: Point-based 3d single stage
  object detector,'' in \emph{Proceedings of the IEEE/CVF conference on
  computer vision and pattern recognition}, 2020, pp. 11\,040--11\,048.

\bibitem{qi2017pointnet++}
C.~R. Qi, L.~Yi, H.~Su, and L.~J. Guibas, ``Pointnet++: Deep hierarchical
  feature learning on point sets in a metric space,'' \emph{arXiv preprint
  arXiv:1706.02413}, 2017.

\bibitem{shi2020points}
S.~Shi, Z.~Wang, J.~Shi, X.~Wang, and H.~Li, ``From points to parts: 3d object
  detection from point cloud with part-aware and part-aggregation network,''
  \emph{IEEE transactions on pattern analysis and machine intelligence}, 2020.

\bibitem{shi2020pv}
S.~Shi, C.~Guo, L.~Jiang, Z.~Wang, J.~Shi, X.~Wang, and H.~Li, ``Pv-rcnn:
  Point-voxel feature set abstraction for 3d object detection,'' in
  \emph{Proceedings of the IEEE/CVF Conference on Computer Vision and Pattern
  Recognition}, 2020, pp. 10\,529--10\,538.

\bibitem{deng2020voxel}
J.~Deng, S.~Shi, P.~Li, W.~Zhou, Y.~Zhang, and H.~Li, ``Voxel r-cnn: Towards
  high performance voxel-based 3d object detection,'' \emph{arXiv preprint
  arXiv:2012.15712}, 2020.

\bibitem{sheng2021improving}
H.~Sheng, S.~Cai, Y.~Liu, B.~Deng, J.~Huang, X.-S. Hua, and M.-J. Zhao,
  ``Improving 3d object detection with channel-wise transformer,'' in
  \emph{Proceedings of the IEEE/CVF International Conference on Computer
  Vision}, 2021, pp. 2743--2752.

\bibitem{vaswani2017attention}
A.~Vaswani, N.~Shazeer, N.~Parmar, J.~Uszkoreit, L.~Jones, A.~N. Gomez,
  {\L}.~Kaiser, and I.~Polosukhin, ``Attention is all you need,'' in
  \emph{Advances in neural information processing systems}, 2017, pp.
  5998--6008.

\bibitem{zhao2020exploring}
H.~Zhao, J.~Jia, and V.~Koltun, ``Exploring self-attention for image
  recognition,'' in \emph{Proceedings of the IEEE/CVF Conference on Computer
  Vision and Pattern Recognition}, 2020, pp. 10\,076--10\,085.

\bibitem{cheng2021masked}
B.~Cheng, I.~Misra, A.~G. Schwing, A.~Kirillov, and R.~Girdhar,
  ``Masked-attention mask transformer for universal image segmentation,''
  \emph{arXiv preprint arXiv:2112.01527}, 2021.

\bibitem{Geiger2012kitti}
A.~Geiger, P.~Lenz, and R.~Urtasun, ``Are we ready for autonomous driving? the
  kitti vision benchmark suite,'' in \emph{Conference on Computer Vision and
  Pattern Recognition (CVPR)}, 2012.

\bibitem{caesar2020nuscenes}
H.~Caesar, V.~Bankiti, A.~H. Lang, S.~Vora, V.~E. Liong, Q.~Xu, A.~Krishnan,
  Y.~Pan, G.~Baldan, and O.~Beijbom, ``nuscenes: A multimodal dataset for
  autonomous driving,'' in \emph{Proceedings of the IEEE/CVF conference on
  computer vision and pattern recognition}, 2020, pp. 11\,621--11\,631.

\bibitem{mao2021voxel}
J.~Mao, Y.~Xue, M.~Niu, H.~Bai, J.~Feng, X.~Liang, H.~Xu, and C.~Xu, ``Voxel
  transformer for 3d object detection,'' in \emph{Proceedings of the IEEE/CVF
  International Conference on Computer Vision}, 2021, pp. 3164--3173.

\bibitem{pan20213d}
X.~Pan, Z.~Xia, S.~Song, L.~E. Li, and G.~Huang, ``3d object detection with
  pointformer,'' in \emph{Proceedings of the IEEE/CVF Conference on Computer
  Vision and Pattern Recognition}, 2021, pp. 7463--7472.

\bibitem{misra2021end}
I.~Misra, R.~Girdhar, and A.~Joulin, ``An end-to-end transformer model for 3d
  object detection,'' in \emph{Proceedings of the IEEE/CVF International
  Conference on Computer Vision}, 2021, pp. 2906--2917.

\bibitem{fan2021embracing}
L.~Fan, Z.~Pang, T.~Zhang, Y.-X. Wang, H.~Zhao, F.~Wang, N.~Wang, and Z.~Zhang,
  ``Embracing single stride 3d object detector with sparse transformer,''
  \emph{arXiv preprint arXiv:2112.06375}, 2021.

\bibitem{liu2019pvcnn}
Z.~Liu, H.~Tang, Y.~Lin, and S.~Han, ``Point-voxel cnn for efficient 3d deep
  learning,'' in \emph{Advances in Neural Information Processing Systems},
  2019.

\bibitem{dosovitskiy2020image}
A.~Dosovitskiy, L.~Beyer, A.~Kolesnikov, D.~Weissenborn, X.~Zhai,
  T.~Unterthiner, M.~Dehghani, M.~Minderer, G.~Heigold, S.~Gelly,
  \emph{et~al.}, ``An image is worth 16x16 words: Transformers for image
  recognition at scale,'' \emph{arXiv preprint arXiv:2010.11929}, 2020.

\bibitem{zeiler2014visualizing}
M.~D. Zeiler and R.~Fergus, ``Visualizing and understanding convolutional
  networks,'' in \emph{European conference on computer vision}.\hskip 1em plus
  0.5em minus 0.4em\relax Springer, 2014, pp. 818--833.

\bibitem{raghu2021vision}
M.~Raghu, T.~Unterthiner, S.~Kornblith, C.~Zhang, and A.~Dosovitskiy, ``Do
  vision transformers see like convolutional neural networks?'' \emph{Advances
  in Neural Information Processing Systems}, vol.~34, 2021.

\bibitem{zhao2021point}
H.~Zhao, L.~Jiang, J.~Jia, P.~H. Torr, and V.~Koltun, ``Point transformer,'' in
  \emph{Proceedings of the IEEE/CVF International Conference on Computer
  Vision}, 2021, pp. 16\,259--16\,268.

\bibitem{lin2017focal}
T.-Y. Lin, P.~Goyal, R.~Girshick, K.~He, and P.~Doll{\'a}r, ``Focal loss for
  dense object detection,'' in \emph{Proceedings of the IEEE international
  conference on computer vision}, 2017, pp. 2980--2988.

\bibitem{he2020structure}
C.~He, H.~Zeng, J.~Huang, X.-S. Hua, and L.~Zhang, ``Structure aware
  single-stage 3d object detection from point cloud,'' in \emph{Proceedings of
  the IEEE/CVF Conference on Computer Vision and Pattern Recognition}, 2020,
  pp. 11\,873--11\,882.

\bibitem{openpcdet2020}
O.~D. Team, ``Openpcdet: An open-source toolbox for 3d object detection from
  point clouds,'' \url{https://github.com/open-mmlab/OpenPCDet}, 2020.

\bibitem{smith2019super}
L.~N. Smith and N.~Topin, ``Super-convergence: Very fast training of neural
  networks using large learning rates,'' in \emph{Artificial Intelligence and
  Machine Learning for Multi-Domain Operations Applications}, vol. 11006.\hskip
  1em plus 0.5em minus 0.4em\relax International Society for Optics and
  Photonics, 2019, p. 1100612.

\bibitem{wang2020infofocus}
J.~Wang, S.~Lan, M.~Gao, and L.~S. Davis, ``Infofocus: 3d object detection for
  autonomous driving with dynamic information modeling,'' in \emph{European
  Conference on Computer Vision}.\hskip 1em plus 0.5em minus 0.4em\relax
  Springer, 2020, pp. 405--420.

\end{thebibliography}

\end{document}